\documentclass[conference]{IEEEtran}
\ifCLASSINFOpdf
\usepackage[pdftex]{graphicx}
\usepackage{algorithm}
\usepackage{algorithmic}

\IEEEoverridecommandlockouts
\usepackage{cite}
\usepackage{bm}
\usepackage{amsmath,amssymb,amsfonts}
\usepackage{algorithmic}
\usepackage{graphicx}
\usepackage{textcomp}
\def\BibTeX{{\rm B\kern-.05em{\sc i\kern-.025em b}\kern-.08em
    T\kern-.1667em\lower.7ex\hbox{E}\kern-.125emX}}
\begin{document}
\title{AMPA-Net: Optimization-Inspired Attention Neural Network for Deep Compressed Sensing\\
}
\author{\IEEEauthorblockN{Nanyu Li$^*$}
\IEEEauthorblockA{\textit{School of Information Engineering and Automation} \\
\textit{Kunming University of science and technology}\\
Kunming, China,650000 \\
email: nanyuli1994@gmail.com}
\and
\IEEEauthorblockN{Charles C. Zhou}
\IEEEauthorblockA{\textit{Quantum Intelligence Inc.}\\
Monterrey, United States, 93943\\
email: charles.zhou@quantumii.com\\
     charleszhou@alum.mit.edu}
}
\maketitle

\begin{abstract}
Compressed sensing (CS) is a challenging problem in image processing due to reconstructing an almost complete image from a limited measurement. To achieve fast and accurate CS reconstruction, we synthesize the advantages of two well-known methods (neural network and optimization algorithm) to propose a novel optimization inspired neural network which dubbed AMP-Net. AMP-Net realizes the fusion of the Approximate Message Passing (AMP) algorithm and neural network. All of its parameters are learned automatically. Furthermore, we propose an AMPA-Net which uses three attention networks to improve the representation ability of AMP-Net. Finally, We demonstrate the effectiveness of AMP-Net and AMPA-Net on four standard CS reconstruction benchmark data sets. Our code is available on https://github.com/puallee/AMPA-Net.
\end{abstract}
\IEEEpeerreviewmaketitle
\section{Introduction}
Compressed sensing (CS) [1-2] theory demonstrates that we can recover a signal from a limited measurement with a high probability if it is sparse in some optimal transform. This new technology is hardware-friendly and widely used in many fields, such as fast magnetic resonance imaging (fMRI) [3], fast and low-dose X-ray imaging [4], radio astronomy imaging [5], single-pixel camera [6] and 3D-video [7]. In this paper, we focus on the CS of natural images, however, our framework can be generalized to other types of data as well.
We show examples of CS in Fig.1. We define that the size of image signal $\bm{X}$ is $\bm{N}_P$, sensing matrix is $\bm{\phi}$, the size of measurement $\bm{Y}$ is $\bm{M}_P$, the degree of under-sampling as CS ratios is $\bm{M}_P$/$\bm{N}_P$, and it also means sampling cost. We reconstruct an almost clear image from the measurement $\bm{Y}$. CS can be written as an optimization for $\bm{l}_1$ norm problem which is presented in Eq. (1):
 \begin{equation}
 \min_{X}\frac{1}{2}\left \| \ Y-\phi X \right \|_{2}^{2}+\lambda\left \| \ D X \right \|_{1}
\end{equation}
 $\bm{\lambda}$ is a regularization parameter for sparsity, and D is the optimal transform. 
\begin{figure}[h]
\centering
\includegraphics[width=0.9\linewidth]{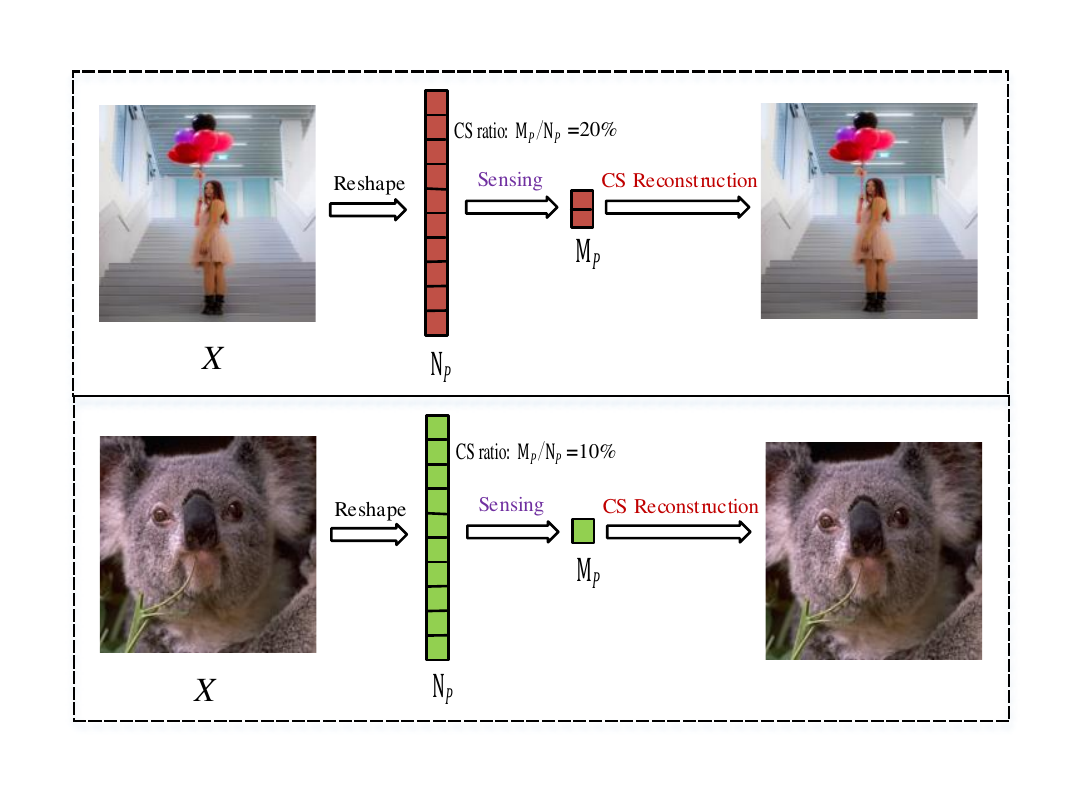} 
\caption{The process of compressed sensing in natural images}
\label{fig:CS} 
\end{figure}
\\Optimization algorithms, neural networks, and optimization-inspired neural networks are mainly used in CS reconstruction. Optimization algorithms: Eq.(1) can be solved by optimization algorithms, such as alternating direction method of multipliers (ADMM) [9], iterative shrinkage-thresholding algorithm (ISTA)[10], and approximate message passing (AMP) [11]. However, these methods are slow to converge. Because natural images are typically non-stationary, it is heavily laboring and time-consuming to design sensing matrix  $\bm{\phi}$ and optimal transform D. Neural network: a fast implicit model for CS Reconstruction, such as DR2-Net [12], Recon-Net [13], Adaptive-Recon-Net [14]. These methods lack accuracy because they do not use prior knowledge of Eq.(1). Optimization-inspired neural network: it achieves the fusion of neural network and optimization algorithm to maintain accuracy and fast speed, such as ADMM-Net [3] fusing ADMM and circular convolution, ISTA-Net$^{+}$ [15] fusing ISTA and convolution neural network. In this paper, we propose a novel optimization-inspired neural network which fuses AMP and neural network, dubbed as AMP-Net, including full connection layer for adaptive sensing, AMP algorithm  and balanced CNN for CS reconstruction. The whole network is trained end to end through Charbonnier loss function. Furthermore, we introduce three attention networks to further improve the performance of CS. 
\\In this paper, Our contributions are four-fold: 
\begin{itemize}
\item We propose an AMP-Net to synthesize the advantages of approximate messaging passing and neural networks. Our method deal with the problem of fast and accurate CS reconstruction.
\item Its enhanced version AMPA-Net use three attention networks to improve CS reconstruction performance.
\item To best of our knowledge, we are the first to propose AMP-Net based on the optimization-inspired neural network and our AMPA-Net is the first to incorporate feature attention for deep compressed sensing.
\item Extensive experiments on four standard benchmark data sets (\textit{i.e.}, Set11, BSD68, BSDS100, and Urban100) show that our methods can reach state-of-the-art performance in CS reconstruction.
\end{itemize}
\section{Related Work}
In this section, we give a brief review of the optimization-inspired neural network, AMP algorithm, and attention mechanism which are most relevant to our method.
\subsection{Optimization-inspired neural network}
The existing optimization-inspired neural networks, including ADMM-Net, ISTA-Net, and ISTA-Net$^{+}$, usually achieve the fusion of an optimization algorithm and a neural network to synthesize the advantages of both of them. This fusion depends on the similarity between the optimization algorithm and the neural network: (1) the calculation of optimization algorithms are usually differentiable, we can use propagated backward [16] to learn their parameters. (2) Iteration of the optimization algorithm is similar to the deep stacking of neural networks [3]. The optimization can unroll into depth structure.
Through fusion, the structure of these networks has some well-defined explanatory. What's more, they can combine the advantages of optimization algorithm and neural network to achieve fast and accurate CS reconstruction.
\subsection{AMP algorithm}
The AMP algorithm is inspired by the message passing in graph theory [11]. Considering some existing errors of CS reconstruction in each iteration in ISTA, AMP adds an Onsager reaction term, which improves the phase transition [17] to get an accurate solution. Furthermore, the AMP algorithm has a strong expansibility. First, AMP combine with a well-defined denoising filter, such as: NLM-AMP [18], BM3D-prgamp [19]. Second, AMP combine with trained denoising filer, such as: LD-AMP [20]. However, there is no real fusion through end to end learning in the above methods.
\subsection{Attention mechanism}
The attention mechanism is an effective method in deep learning, such as SE-Net[22] in image classification, RCAN [23]  in image super-resolution. In CS, we introduce three attention networks to redistribute spatial and channel information of our AMP-Net.
\section{AMP-Net}
The overview structure of AMP-Net is illustrated in Fig.2. Its structure can be corresponding to Algorithm.2.  We will describe our approach in detail from three aspects: Algorithm, Framework, and  Loss function.
\begin{figure*}[t]
\centering
\includegraphics[width=\linewidth]{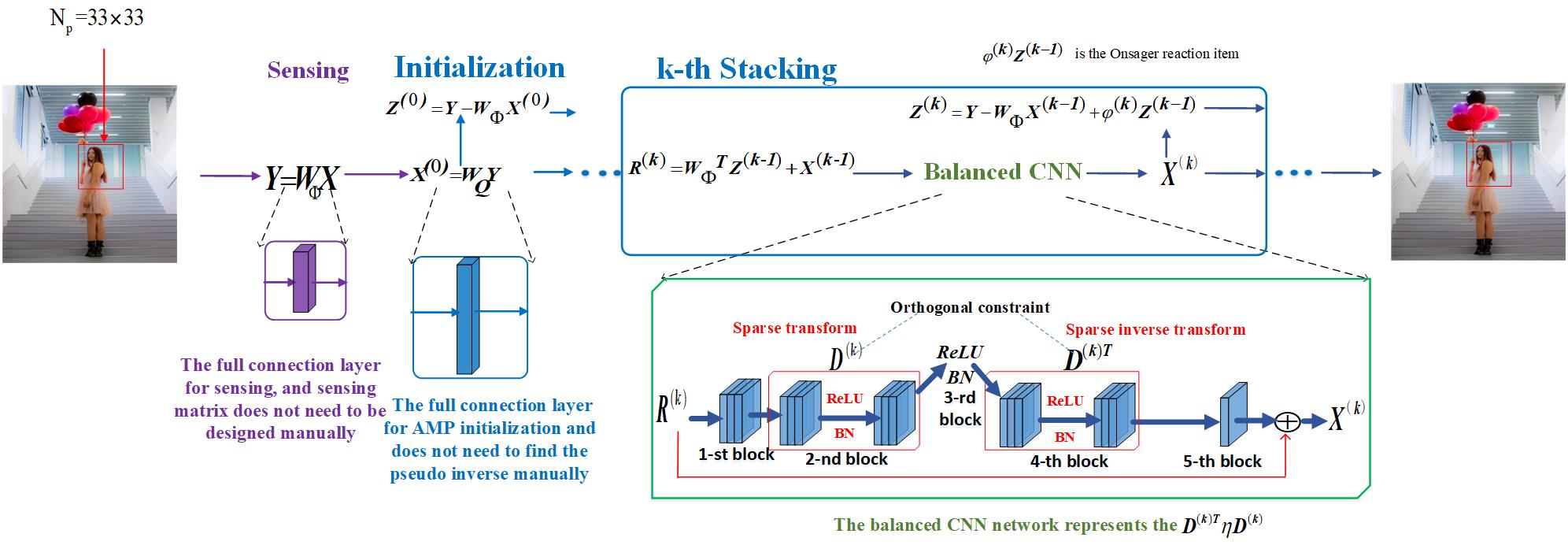} 
\caption{The diagram of proposed AMP-Net,parts of the formulas is amp algorithm, the others are neural network}
\label{fig:AMPNet} 
\end{figure*}
\subsection{Algorithm}
David[11] proposed an AMP algorithm to solve Eq.(1), AMP algorithm is presented in Algorithm.1.
\begin{algorithm}
\caption{Approximate Message Passing}
\label{alg:Approximate Message Passing}
\begin{algorithmic}
\REQUIRE $Y, \phi$
\ENSURE $X=D^TS^{(N)}$
\\$Parameters:m$ 
\\Initialization
\STATE {$S^{(0)}=pinv(\phi)Y, 
Z^{(0)}=Y- \phi S^{(0)}$} 
\\While not converge,do:
       \STATE $  S^{(k)}=\eta(\phi^T Z^{(k-1)}+S^{(k-1)})$
       \STATE $  Z^((k) )=Y- \phi S^{(k-1)}+ \varphi^{(k)} Z^{(k-1)}$
        \STATE $  \varphi^{(k)}=m(\eta{'}(\phi^{T} Z^{(k-1)}+S^{(k-1)}))$
        \\

\end{algorithmic}
\end{algorithm}
In Algorithm.1, the input is measurement $\bm{Y}$ and handcrafted sensing matrix $\bm{\phi}$. The output is reconstructed image $\bm{X}$.
AMP includes initialization and iteration. In initialization, $\bm{D}$ is handcrafted optimal transform, $\bm S^{(0)}$ is reconstructed sparse representation[21]. $\bm{pinv}$ is a pseudo-inverse operation. 
In iteration, $\bm{k}$ is an iteration index,$\bm Z^{(k)}$ is current residual. ${\eta}$ is threshold function for  $\bm{l}_1$  regulation, and $\bm{\eta{'}}$ is its derivative of function input. $\bm{m}$ and  $\bm{\varphi^{(k)}}$ are scale factors. $\bm{\varphi^{(k)} Z^{(k-1)}}$ is the Onsager reaction item to improve phase transition. In their experiments, AMP needs almost one hundred of iteration to converge for neural images. 

\subsection{Framework}
In this part, we expand the traditional AMP to achieve fusion of AMP and neural network. For AMP, when $\bm{D^{(k)}}$ exists orthogonality [25] ($\bm{D^{(k)^T}D^{(k)}=I}$), AMP can be transformed into Eq.(2):
 \begin{equation}
X^{(k)}={D^{(k)}}^{T}\eta D^{(k)}(\phi^T Z^{(k-1)}+X^{(k-1)}) 
\end{equation}
In Eq.(2), $\bm{D^{(k)}}$ means that D can be different in each iteration. Furthermore, we design a balanced CNN to represent $\bm{{D^{(k)}}^{T}\eta D^{(k)}}$, and this balanced CNN is guided by the AMP algorithm to select the optimal solution $\bm{X^{(k)}}$, which is presented in Eq.(3):
\begin{equation}
X^{(k)}=f^{(k)}(\phi^T Z^{(k-1)}+X^{(k-1)})  
\end{equation}
$\bm{f^{(k)}}$ is a balanced CNN, it’s parameters is $\bm{P^{(k) }}$. After that, We also use full-connection layer network for adaptive sensing[14] and initialization to replace handcrafted metric $\bm{\phi}$ and $\bm{pinv}$ calculation.

\begin{algorithm}
\caption{Approximate Message Passing-Inspired Neural Network}
\label{alg:Approximate Message Passing-Inspired Neural Network}
\begin{algorithmic}
\REQUIRE $Y= W_\phi X, W_\phi$
\ENSURE $X^{(N)}$
\\Parameters: $\bm{\theta=(W_Q,W_\phi,P^{(k)},\varphi^{(k)})}$, 
\\Initialization
\STATE $X^{(0)}=W_QY$, 
\STATE $Z^{(0)}=Y-W_\phi X^{(0)}$,
\\For k=1to N:  
       \STATE $  R^{(k)}=W_\phi^T Z^{(k-1)}+X^{(k-1)}$
       \STATE $  X^{(k)}=f^{(k)} ( R^{(k)})$
        \STATE $ Z^{(k)}=Y- W_\phi X^{(k-1)}+  \varphi^{(k)} Z^{(k-1) }$
        \\

\end{algorithmic}
\end{algorithm}
In Fig.2, arrows direct the flow of AMP-Net. $\bm{ W_\phi}$ is learning sense matrix, and  $\bm{ W_Q}$ is learning initialization.
we define K iteration as K stacking: (we use K = 9, the traditional AMP usually requires K$\bm{>}$100), each stacking include AMP calculation and balanced CNN. Inspired by[15], Our balanced CNN contains 5 blocks. Intermediate reconstruction $\bm{ R^{(k)}}$ is input, optimal reconstruction $\bm{ X^{(k)}}$ is output. The 1-st block which is composed of a convolution with $\bm{3\times3\times32}$ filters. The 2-nd block as $\bm{ D^{(k)}}$ which is composed of a convolution with $\bm{3\times3\times32}$ filters, batch normalization (BN), Rectified linear unit (ReLU), a convolution with $\bm{3\times3\times32}$filters. The 3-rd block which is composed of batch normalization (BN) and Rectified linear unit (ReLU), which corresponds to the threshold function $\eta$ of AMP because of its non-negativity. The 4-th block as  $\bm{ {D^{(k)}}^T}$ which is composed of a convolution with $\bm{3\times3\times32}$ filters, batch normalization (BN), Rectified linear unit (ReLU), a convolution with $\bm{3\times3\times32}$ filters. The 5-th block which is composed of a convolution with $\bm{3\times3\times1}$ filters. The skip connections (red line) can facilitate training. 

\subsection{Loss function}
Following [13], we use image block as network input to maintain speed and stability, such as in Fig.2, the dimension of blocks is $\bm{33\times33}$. Given the training data $\bm{(X_i)_i^B}$ of each batch size, B is the number of total image block  $\bm{X_i}$. To learn the network parameters $\bm{\theta}$ in the Algorithm.2, we should not only minimize the CS reconstruction error ($\bm{X_i^{(N)}-X_i}$), but also consider the orthogonality ($\bm{D^{(k)^T}D^{(k)}=I)}$, as shown in Eq.(4).
\begin{equation}
L_{total}= L_R+\lambda L_O  
\end{equation}
The loss function includes orthogonal constraints $\bm{L_O}$, and reconstruction errors $\bm{L_R}$, $\bm{\lambda}$ is the regularization parameter, we empirically set $\bm{\lambda}$  to 0.01.
\begin{equation}
 L_R=1/B \sum_{i=1}^B \sqrt{(X_i^{(N)}-X_i)^2+\epsilon^2}
\end{equation}
\begin{equation}
 L_O= 1/B\sum_{i=1}^B\sum_{k=1}^N\sqrt{(D^{(k)^T}D^{(k)}-I)^2+\epsilon^2}
\end{equation}
As shown in Eqs.(5-6), Charbonnier penalty function is adopted for $\bm{L_R}$ and $\bm{L_O}$ with empirically set $\bm{\epsilon}$ to 1e-3.
\section{AMPA-Net}
In this section, the overview structure of AMPA-Net is illustrated in Fig.3, and its structure can be corresponding to Algorithm.3. We will describe our approach in detail from two aspects: Framework and  Attention.
\begin{figure*}[t]
\centering
\includegraphics[width=\linewidth]{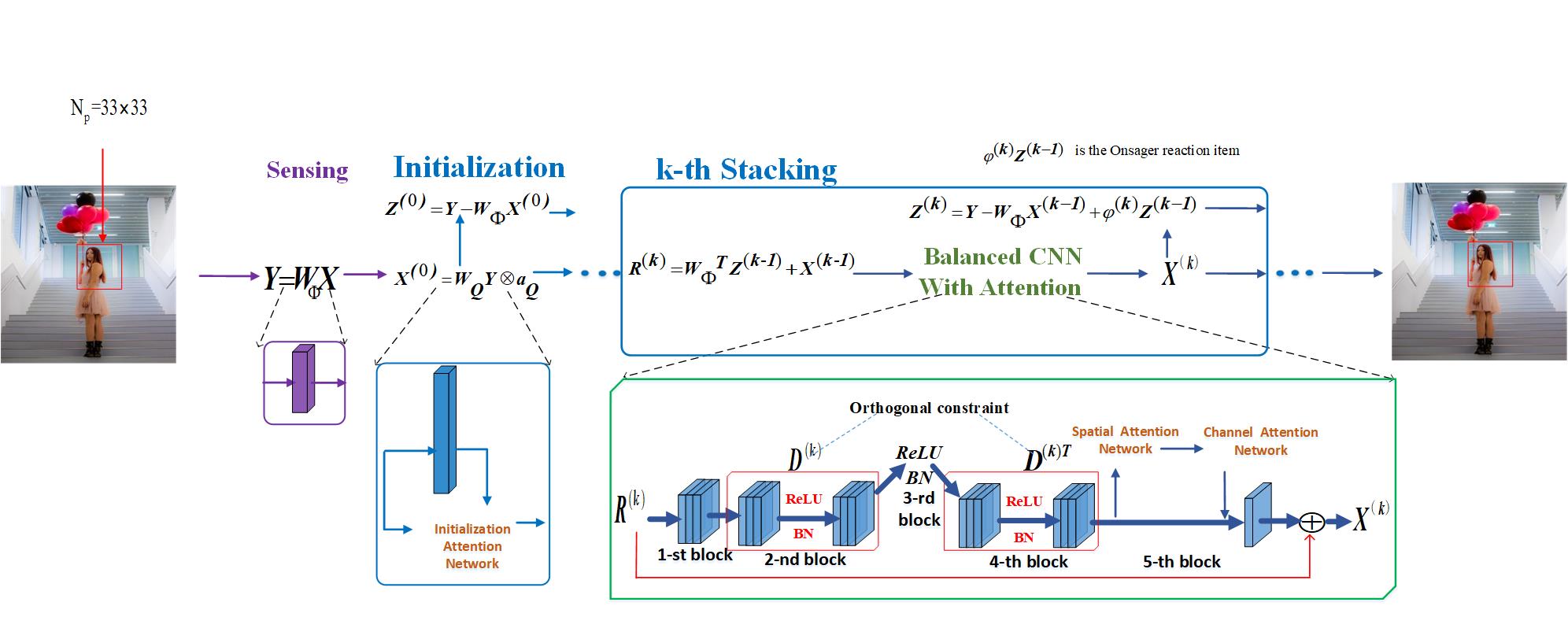} 
\caption{The diagram of proposed AMPA-Net,}
\label{fig:AMPANet} 
\end{figure*}

\begin{algorithm}
\caption{Approximate Message Passing-Inspired Attention Neural Network}
\label{alg:Approximate Message Passing-Inspired Attention Neural Network}
\begin{algorithmic}
\REQUIRE $Y= W_\phi X, W_\phi$
\ENSURE $X^{(N)}$
\\Parameters: $\bm{\theta=(W_Q,W_\phi,P^{(k)}, \varphi^{(k)},a_Q , a_s^{(k)}, a_c^{(k)})}$, 
\\Initialization
\STATE $X^{(0)}=W_QY\otimes a_Q$, 
\STATE $Z^{(0)}=Y-W_\phi X^{(0)}$,
\\For k=1to N:
       \STATE $  R^{(k)}=W_\phi^T Z^{(k-1)}+X^{(k-1)}$
       \STATE $  X^{(k)}=f^{(k)} ( R^{(k)})$
        \STATE $  Z^{(k)}=Y- W_\phi X^{(k-1)}+  \varphi^{(k)} Z^{(k-1) }$
\end{algorithmic}
\end{algorithm}

\subsection{Framework}
In Fig.3, the arrow indicates the flow direction of AMPA-Net. We add three more attention networks for AMP-Net: initialization attention network, spatial attention network, and channel attention network. The loss function of AMPA-Net is the same as AMP-Net, as shown in Eq.(4). Comparing with AMP-Net, the learning parameters of network $\bm{(a_Q , a_s^{(k)}, a_c^{(k)})}$ are added, and others are the same. 
\subsection{Attention}
As mentioned above, we introduce three attention networks for CS reconstruction. These networks are as shown in Fig.4.
\begin{figure}[h]
\centering
\includegraphics[width=0.9\linewidth]{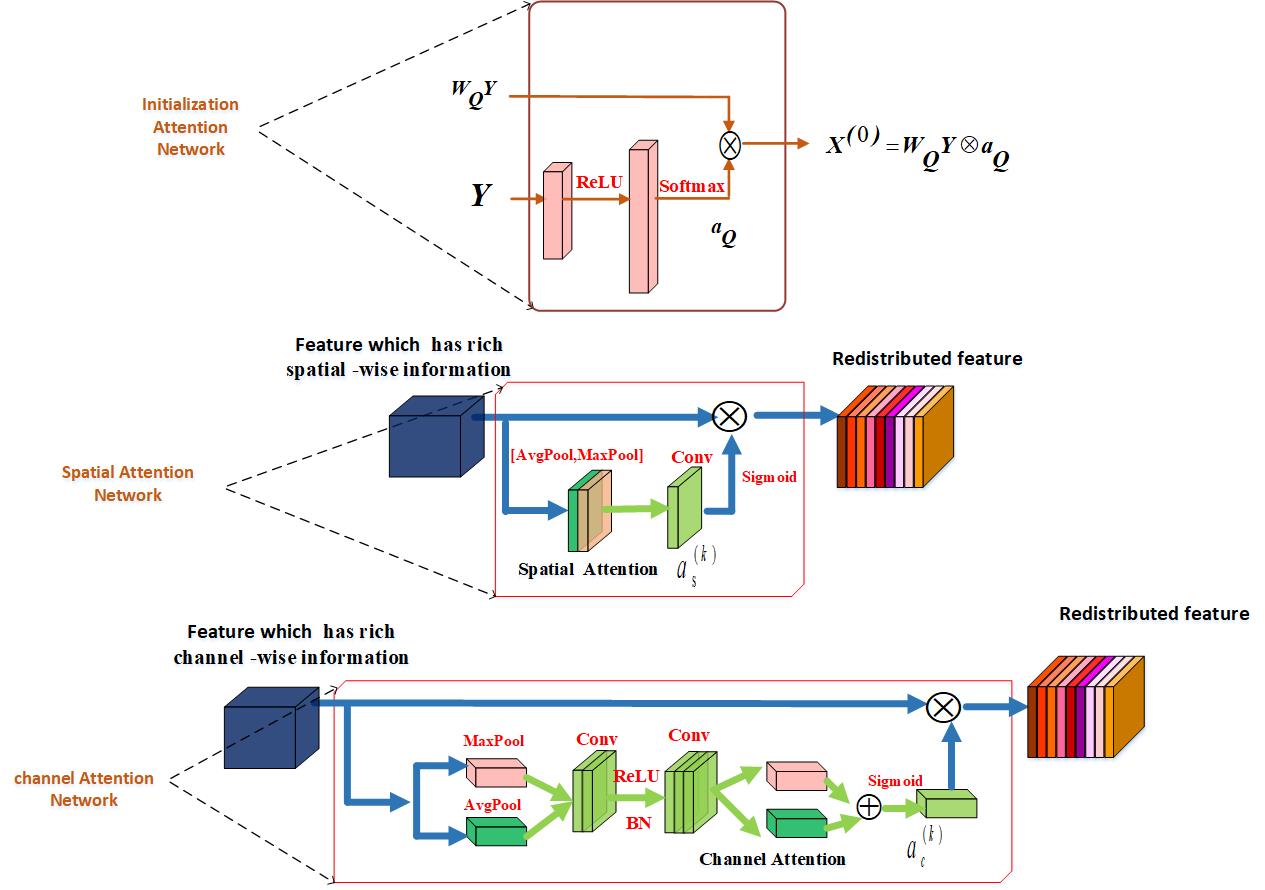} 
\caption{ The diagram of attention networks in AMPA-Net}
\label{fig:3attention} 
\end{figure}

For the initialization attention network, we use a multi-layer perceptron (MLP)  to obtain the attention weight $\bm{a_Q}$. The MLP is composed of a fully connected layer, Rectified linear unit (ReLU), fully connected layer, and softmax function (Softmax). 

For spatial attention networks,enlightened by CBAM [26], we use both global max-pooling and global average-pooling to aggregate global information. First, global information is aggregated through average-pooling and max-pooling respectively and concatenated. Second, a convolution layer is used to get the spatial attention weight  $\bm{a_s^{(k)}}$ which is composed of convolution with $\bm{3\times3\times1}$ filters (Conv) and sigmoid activation function (Sigmoid). Finally, the more informative representation is obtained by element-wise multiplication. 

For channel attention network, enlightened by CBAM [26]: First, global information is aggregated by average-pooling and max-pooling respectively. Second, both of them are forwarded to the same Full Convolution Network, which is composed of convolution with $\bm{3\times3\times8}$ filters (Conv)(with Zero padding), BN (batch normalization), Rectified linear unit (ReLU), convolution with $\bm{3\times3\times32}$ filters(Conv)(with Zero padding), and we add two outputs and  get the channel attention weight $\bm{a_c^{(k)}}$ through sigmoid function (Sigmoid). Finally, the more informative representation is obtained by element-wise multiplication.

\section{ EXPERIMENT}
In this section, we are divided into four parts: experimental setting, hyperparameters, ablation study, experiments in natural image.
\subsection{ Experiment setting}
\begin{bfseries} 
Network setting:
\end{bfseries}
 We use TensorFlow to implement our methods. In detailed, we use Adam optimization with a learning rate 0.0001, batch size of 64, the model's stacking number of 9, and the regularization parameter $\bm{\lambda}$ of 0.01. Empirically, we use Xavier initialization for balanced CNN $\bm{(P^{(k)},a_s^{(k)}, a_c^{(k)})}$ , Gaussian initialization for for $\bm{(W_Q,W_\phi, a_Q)}$  and the 0.1 initialization for $\bm{\varphi^{(k)}}$. All models are trained and tested on Linux with GTX 1080ti GPUs.
 
\begin{bfseries} 
Training setting:
\end{bfseries} 
 For the fair comparison, we use the same 91 images as the training set which has been used in previous CS works [12-15]:  random extract luminance component of 8912 randomly cropped image block (each of size $\bm{33\times33}$) of the image set. According to different CS ratios: $\bm{(1\%,4\%,10\% ,25\%.30\%,40\%,50\%)}$, we train separately AMP-Net and AMP-Net.

\begin{bfseries} 
Testing setting:
\end{bfseries} 
Following [12-15], we use two standard benchmark datasets for testing in CS: Set 11[12], BSD68[32], which have 11 and 68 gray images respectively. To test the generalization ability of our methods in a larger dataset or multichannel dataset, we test RGB image datasets (using the same network to recover the individual channels): Urban100 [28], BSDS100 [29], which have 100 and 100 RGB images.

\subsection{Hyperparameters}

We study the impact of varying hyperparameters (epoch, stacking, batch-size, active function, the regularization parameter $\gamma$, loss function) of AMP-Net and AMPA-Net,  and CS ratio is {25\%}, the test set is Set11 because they are often used [15]. 
First, we study the impact of epoch and stacking numbers of our methods, other hyperparameters follow the previous network setting. Fig.5.(a) shows the average Peak Signal to Noise Ratio(PSNR) curves for Set11 of different numbers of the epoch, Fig.5.(b) shows the average PSNR curves for Set11 of different numbers of stacking. Because of the trade-off between complexity and performance, it can be seen from Fig.5 that epoch 200 and stacking 9 are the most appropriate parameters of our networks.
\begin{figure}[h]
\centering
\includegraphics[width=0.9\linewidth]{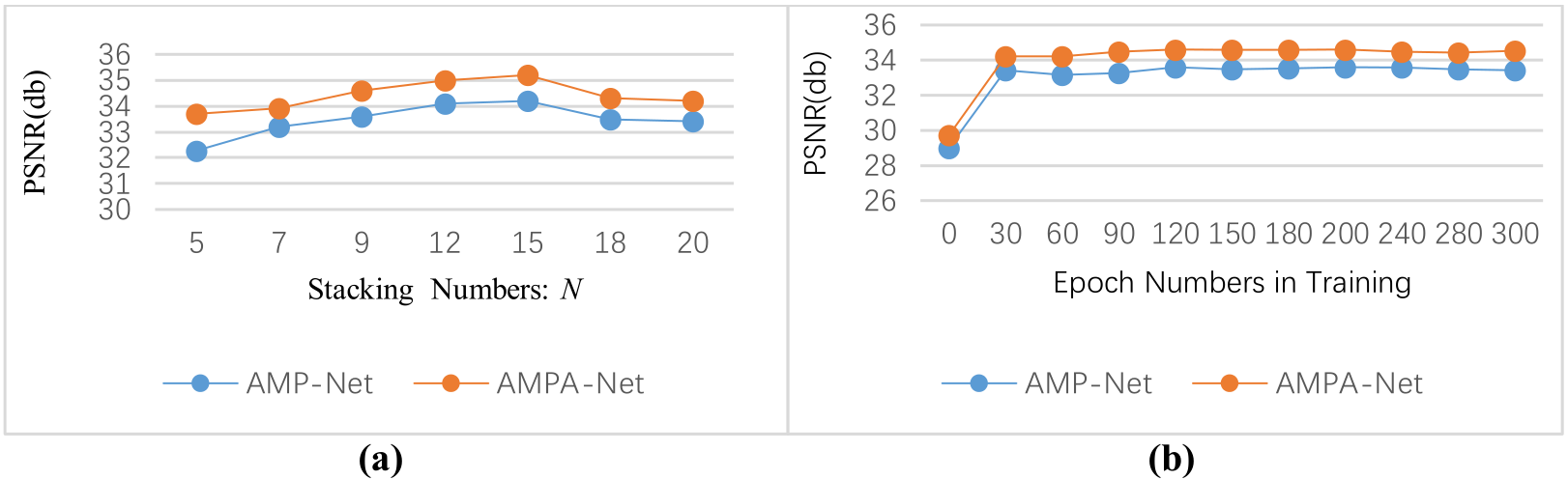} 
\caption{The average PSNR curves for with respect to different numbers of stacking and epoch}
\label{T6} 
\end{figure}

Second, we also study the impact of batch-size and active function of AMP-Net and AMPA-Net, other hyperparameters also follow the previous network setting. Fig.6.(a) shows the average PSNR curves for Set11 of different numbers of batch-size, Fig.6.(b) shows the average PSNR curves for Set11 of different types of activation functions. They show that batch-size 64 and active function ReLU are more suitable for our networks.
\begin{figure}[h]
\centering
\includegraphics[width=0.9\linewidth]{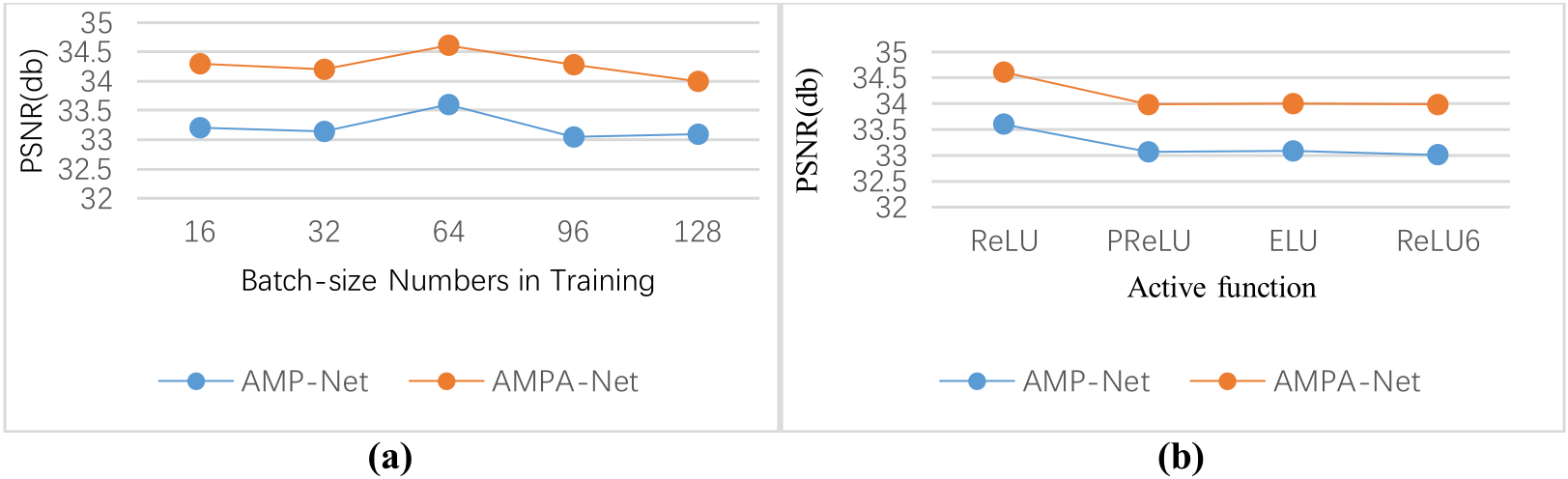} 
\caption{The average PSNR curves for with respect to different batch-size and active function}
\label{T7} 
\end{figure}
Finally, we also study the impact of the regularization parameter $\bm{\lambda}$ and loss function and emphasize that other hyperparameters also follow the previous network setting. Fig.7.(a) shows the average PSNR curves for Set11 of different numbers of  $\bm{\lambda}$, Fig.7.(b) shows the average PSNR curves for Set11 of different types of loss functions.
\begin{figure}[h]
\centering
\includegraphics[width=0.9\linewidth]{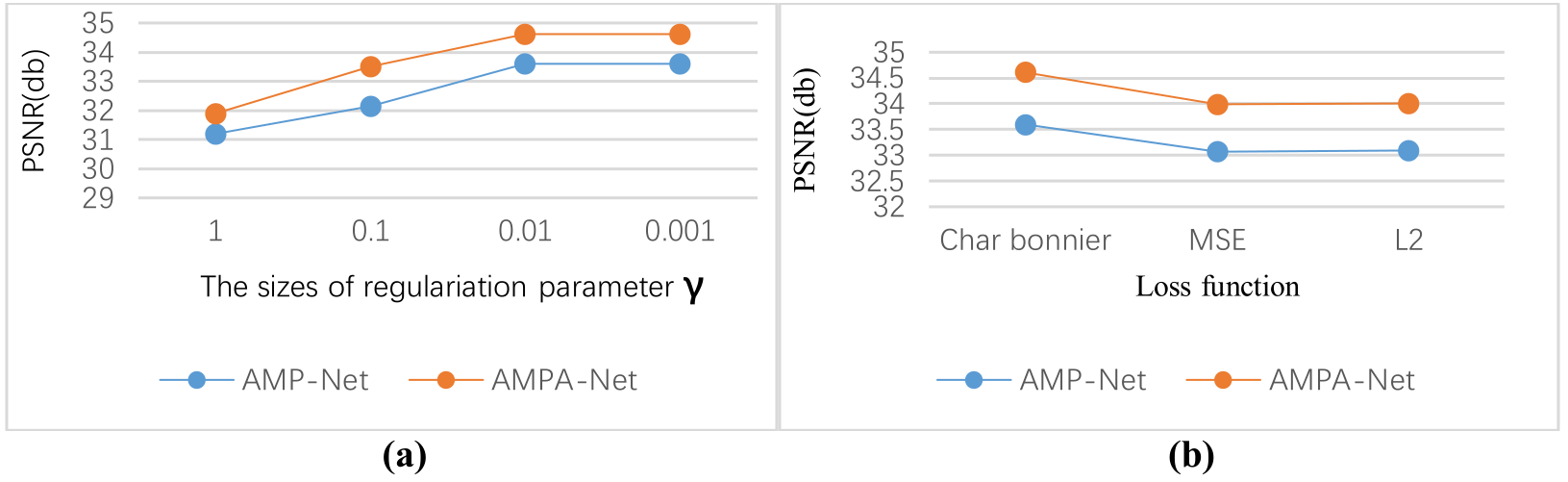} 
\caption{The average PSNR curves for with respect to different the regularization parameter $\gamma$ and loss function}
\label{F3} 
\end{figure}
\subsection{Ablation studies}
In this sub-section, we study the contribution of AMP, balanced CNN ,the sensing network $\bm{(W_\phi)}$ in AMP-Net,initialization attention$\bm{(a_Q)}$, spatial attention $\bm{(a_c^{(k)})}$ and channel attention $\bm{(a_s^{(k)})}$ in AMPA-Net, to analyze the effectiveness of each component in our method.
For testing, benchmark data is Set11[13], CS ratios are $\bm{(10\%,30\%,50\%)}$. The results are shown in Tab.1. 
In Tab.1, compared with single balanced CNN,  single AMP is more important on the accuracy of CS reconstruction.
Compared with the random Gaussian matrix, $\bm{W_\phi}$  can extract more effective information to assist CS reconstruction[30]. The effect of attention mechanism on AMP-Net is more obvious when the CS rate is higher.
\begin{table}[ht]
\centering
\caption{Contribution of attention and their combinations in Set11}
\includegraphics[width=0.9\linewidth]{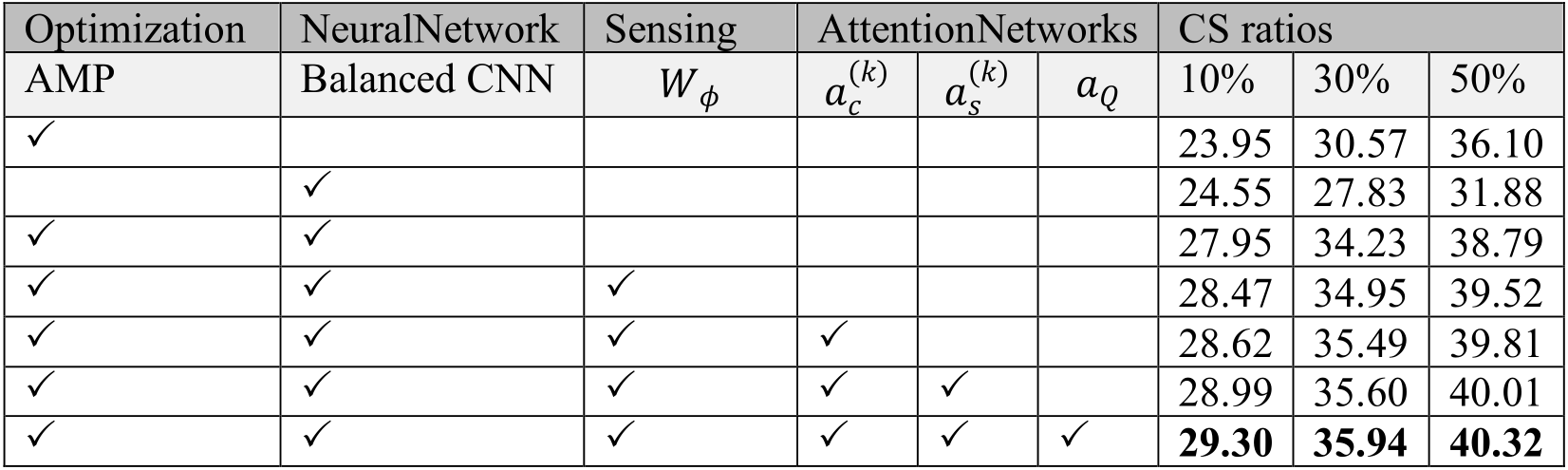}
\label{tab:T1}
\end{table}
\subsection{Experiments in natural image}
We compared our proposed AMP-Net, AMPA-Net with the state-of-the-art image CS methods including optimization methods: BM3D-AMP [19], LD-AMP [20]; neural networks: DR2-Net[12], Adaptive-Recon-Net (Ad-Recon-Net)[14], FCMN[32], Full-Conv[33]; optimization-inspired neural networks: ISTA-Net, ISTA-Net$\bm^{+}$[15].  CS ratios are $\bm{(1\%,4\%,10\% ,25\%,40\%,50\%)}$, and the sensing matrix is random Gaussian under-sampling.
\begin{table}[ht]
\centering
\caption{Average PSNR (dB) and speed performance comparisons on Set 11}
\includegraphics[width=0.9\linewidth]{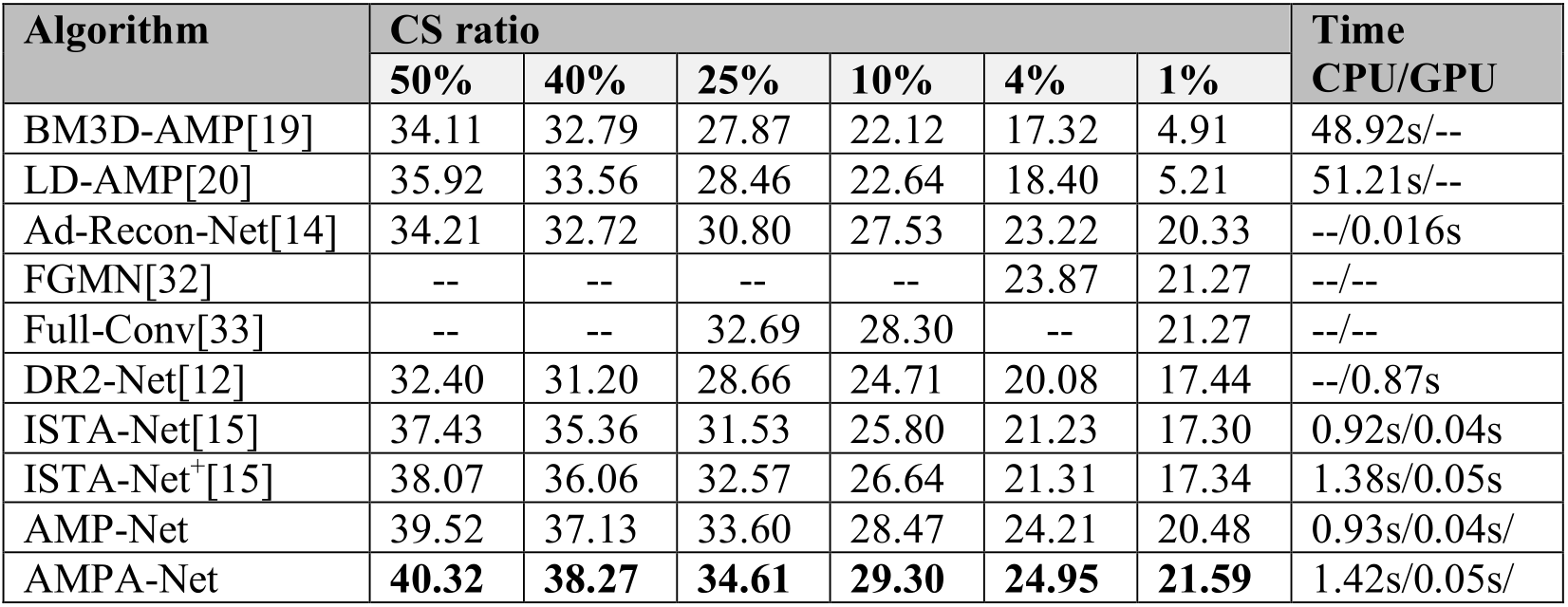}
\label{tab:T2}
\end{table}

\begin{figure}[ht]
\centering
\caption{Average PSNR (dB) performance comparisons on BSD68}
\includegraphics[width=0.9\linewidth]{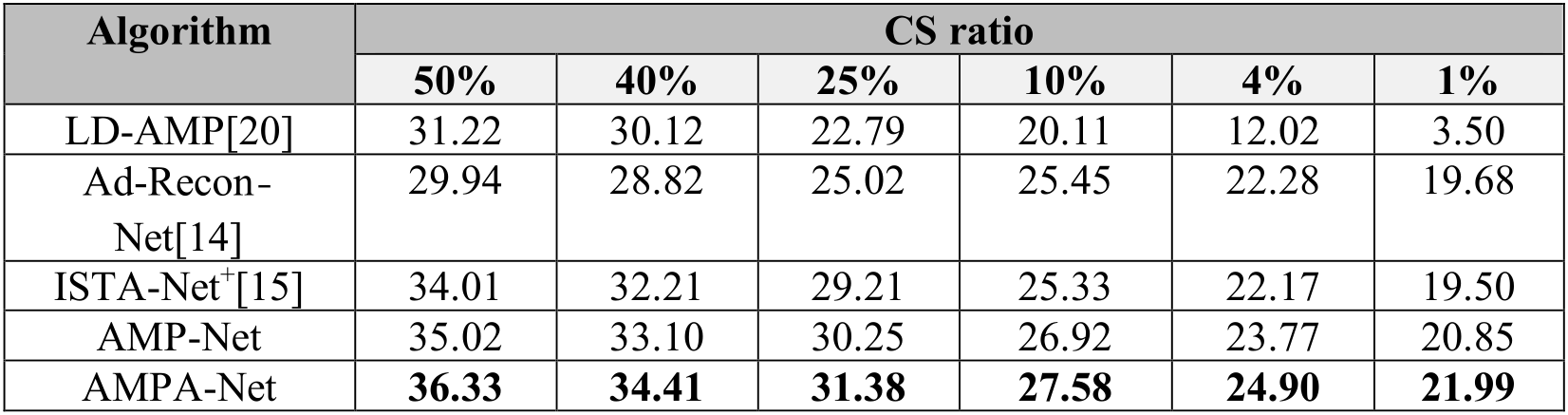}
\label{tab:T3}
\end{figure}

\begin{figure}[ht]
\centering
\caption{Average PSNR (dB) performance comparisons on BSDS100}
\includegraphics[width=0.9\linewidth]{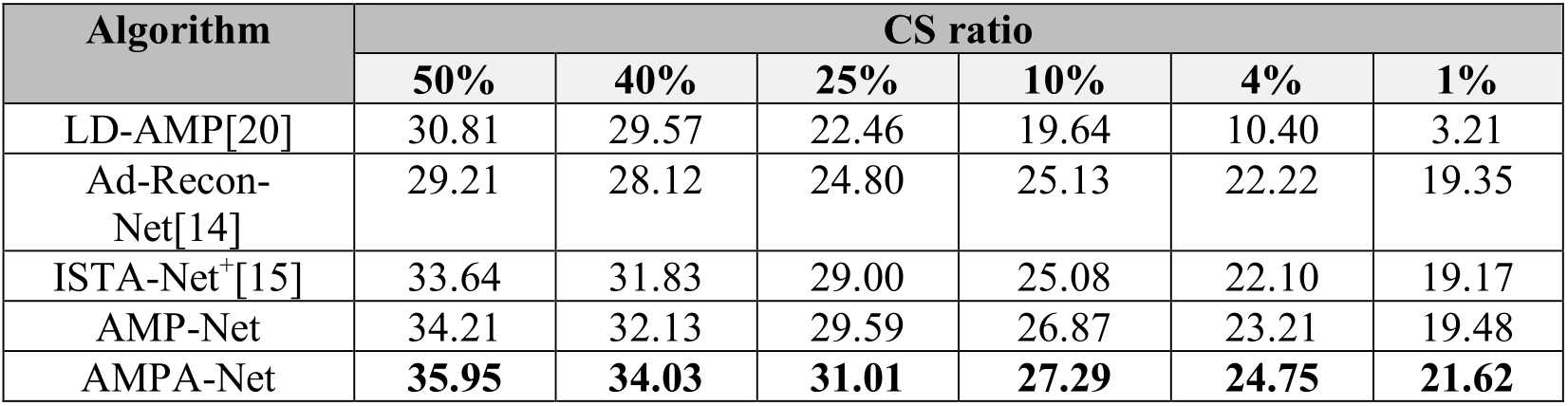}
\label{tab:T4}
\end{figure}

\begin{figure}[ht]
\centering
\caption{Average PSNR(dB) performance comparisons on Urban100}
\includegraphics[width=0.9\linewidth]{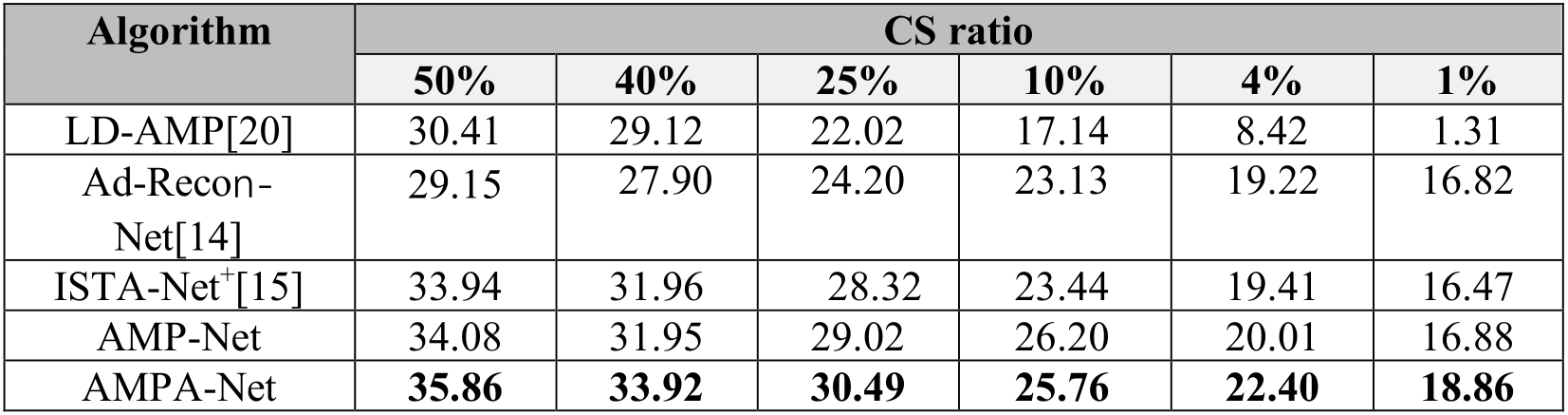}
\label{tab:T5}
\end{figure}
Tab.2 shows the results of all methods on Set11, including reconstruction accuracy and average time of reconstructing per-image. We observed that ISTA-Net, and ISTA-Net$\bm^{+}$ benefited from the optimization-inspired design, achieving high accuracy across all defined CS ratios, as well as maintaining fast speed. The optimization algorithms achieve high accuracy at high CS ratios but maintain low speed, especially LD-AMP. Neural network algorithms achieve well accuracy at low CS ratios as well as maintaining high speed, especially Adaptive Recon-Net. However, our proposed AMP-Net and AMPA-Net almost outperform all the existing methods across all defined CS ratios and also maintain fast speed. Validating the generalizability of our proposed network on larger datasets (BSD68, BSDS100, and Urban100). In Tab.3-5, we compared our models with the state-of-the-art methods: LD-AMP, Ad-Recon-Net, ISTA-Net, and ISTA-Net$\bm^{+}$, and our models still surpasses other methods under all defined CS ratios.
\section{Conclusion}
In this paper, our method have made three highlights: (1) We propose an AMP-Net which achieve the fusion of AMP algorithm and balanced CNN to address the problem of fast and accurate CS reconstruction. (2) We propose the AMPA-Net which uses three attention networks to improve our AMP-Net. (3) Extensive experiments on four CS reconstruction benchmark data sets verify the effectiveness of our AMP-Net and AMPA-Net, including the convergence, speed, accuracy of the models, and the contributions of neural network, optimization algorithm, attention module. Since our AMP-Net and AMPA-Net are quite efficient, one direction of interest for our future work is expanding them to deal with other optimization problems and perform a more theoretical analysis of them.
\section{Acknowledge}
We are committed to develop innovative compressed sensing technology in this paper. This technology(AMP-Net and AMPA-Net) has been deployed in Kunming University of science and technology(KUST) in March 2019, and the patent is filed in March 2020 as well. Our project is fully supported by Quantum Intelligence inc. in Monterrey, United States.


\begin{thebibliography}{1}
\bibitem{ref1}Donoho DL. Compressed sensing. IEEE Trans. on Information Theory, vol.52, no.4, pp.1289-1306, 2006.  
\bibitem{ref2}Candes E J, Tao T. Near-Optimal Signal Recovery From Random Projections: Universal Encoding Strategies. IEEE Trans. on Information Theory, vol.52,no.12, pp.5406-5425, 2006.
\bibitem{ref3}J Sun, H Li, Z Xu. Deep ADMM-Net for compressive sensing MRI. In Advances in neural information processing systems (NIPS), pp.10-18, 2016.
\bibitem{ref4}Yoseop Han, Jaejoon Yoo, Jong Chul Ye: Deep Residual Learning for Compressed Sensing CT Reconstruction via Persistent Homology Analysis. arXiv preprint arXiv: 1611.06391, 2016  
\bibitem{ref5}Wenger S, Magnor M, Y. Pihlström. SparseRI: A Compressed Sensing Framework for Aperture Synthesis Imaging in Radio Astronomy. Publications of the Astronomical Society of the Pacific, vol.122, no.897, pp.1367-1374, 2010.
\bibitem{ref6}Duarte M F, Davenport M A , Takhar D. Single-Pixel Imaging via Compressive Sampling. IEEE Signal Processing Magazine, vol.25, no.2, pp.83-91, 2008.
\bibitem{ref7}Edgar M P, Gibson G M, Spalding G Cl. Real-time 3D video utilizing a compressed sensing time-of-flight single-pixel camera. Optical Trapping \& Optical Micromanipulation XIII. Optical Trapping and Optical Micromanipulation XIII, pp.9921-99222, 2016.
\bibitem{ref8}Ruderman D L, Statistics of natural images. Network Computation in Neural Systems, vol.5, no.4, pp.517-548, 1994.
\bibitem{ref9}Boyd S, Parikh N, Chu E. Distributed Optimization and Statistical Learning via the Alternating Direction Method of Multipliers. Foundations \& Trends in Machine Learning, vol.3, no.1, pp.1-122, 2010.
\bibitem{ref10}Beck A, Teboulle M. A Fast Iterative Shrinkage-Thresholding Algorithm for Linear Inverse Problems. SIAM Journal on Imaging Sciences, vol.2, no.1, pp.183-202, 2009.
\bibitem{ref11}David L. Donoho, Arian Maleki, Andrea Montanari. Message-passing algorithms for compressed sensing. In Proceedings of the National Academy of Sciences of the United States of America, vol.106, no.45, pp.18914, 2009.
\bibitem{ref12}H Yao, F Dai, D Zhang, Y Ma, S Zhang, Y Zhang. DR2-Net: Deep Residual Reconstruction Network for Image Compressive Sensing. arXiv preprint arXiv: 1702.05743, 2017 
\bibitem{ref13}Kulkarni K, Lohit S, Turaga P. ReconNet: Non-Iterative Reconstruction of Images from Compressively Sensed Random Measurements. In Proceedings of IEEE Conference on Computer Vision and Pattern Recognition (CVPR), pp.449-458, 2016.
\bibitem{ref14}Xie X, Wang Y, Shi G. Adaptive Measurement Network for CS Image Reconstruction. In Proceedings of CCF Chinese Conference on Computer Vision (CCCV). Springer, Singapore, pp.407-417, 2017.
\bibitem{ref15}Zhang J, Ghanem B. ISTA-Net: Iterative Shrinkage-Thresholding Algorithm Inspired Deep Network for Image Compressive Sensing. In Proceedings of IEEE Conference on Computer Vision and Pattern Recognition (CVPR), 2018.
\bibitem{ref16}Vogl T P,  Mangis J K,  Rigler A K. Accelerating the convergence of the back-propagation method. Biological Cybernetics, vol.59, no.4-5, pp.257-263, 1998.
\bibitem{ref17}Maleki A, Montanari A. Analysis of approximate message passing algorithm. In IEEE Information Sciences \& Systems, pp.1-7,2010.
\bibitem{ref18}Metzler, Christopher A., Arian Maleki, and Richard G. Baraniuk. From denoising to compressed sensing. IEEE Trans. on Information Theory, vol.62, no.9, pp.5117-5144, 2016
\bibitem{ref19}Metzler, Christopher A., Arian Maleki, and Richard G. Baraniuk. BM3D-prgamp: compressive stacking retrieval based on BM3D denoising. In Proceedings of IEEE International Conference on Image Processing (ICIP), pp.2504-2508, 2016
\bibitem{ref20}Metzler, Christopher A., Ali Mousavi, and Richard Baraniuk. Learned D-AMP: Principled Neural Network based Compressive Image Recovery. In Advances in Neural Information Processing Systems (NIPS), pp.1772-1783, 2017
\bibitem{ref21}Nanyu Li,Yujuan Si,Duo Deng,Chunyu,Yuan.ECG beats classification via online sparse dictionary and time pyramid matching,In Proceedings of IEEE 20th International Conference on Communication Technology (ICCT), pp.1537-1543, 2017.
\bibitem{ref22}Hu J, Shen L, Sun G. Squeeze-and-Excitation Networks. In Proceedings of IEEE Conference on Computer Vision and Pattern Recognition (CVPR), pp.7132-7141, 2018.
\bibitem{ref23}Zhang Y, Li K, Li K, Lichen W, Bineng Z, Yun F. Image Super-Resolution Using Very Deep Residual Channel Attention Networks. In Proceedings of the European Conference on Computer Vision (ECCV), pp.286-301, 2018.
\bibitem{ref24}Zhang J, Xia L, Huang M, et al. Image reconstruction in Compressed Sensing based on single-level DWT, In Proceedings of IEEE Workshop on Electronics, Computer \& Applications. pp.941-944, 2014.
\bibitem{ref25}Liu Y, Zhan Z, Cai J, Xiaobo,Q. Projected Iterative Soft-thresholding Algorithm for Tight Frames in Compressed Sensing Magnetic Resonance Imaging. IEEE Trans on Medical Imaging, vol.35, no.9, pp.2130-2140, 2016.
\bibitem{ref26}Woo S, Park J, Lee J Y, et al. CBAM: Convolutional Block Attention Module. In Proceedings of the European Conference on Computer Vision (ECCV), pp.3-19, 2018.
\bibitem{ref27}Nelson J, Price E, Wootters M. New constructions of RIP matrices with fast multiplication and fewer rows. In Proceedings of the twenty-fifth annual ACM-SIAM symposium on Discrete algorithms, pp.1515-1528, 2012.
\bibitem{ref28}Huang, J.B., Singh, A., Ahuja, N. Single image super-resolution from transformed self-exemplars. In Proceedings of IEEE Conference on Computer Vision and Pattern Recognition (CVPR), pp.5197-5206, 2015.
\bibitem{ref29}Martin D, Fowlkes C, Tal D. A database of human segmented natural images and its application to evaluating segmentation algorithms and measuring ecological statistics. In Proceedings of International Conference on Computer Vision (ICCV), 2001.
\bibitem{ref30}Yan Wu, Mihaela Rosca, Timothy Lillicrap: deep compressed sensing. In Proceedings of the International Conference on Machine Learning (ICML), 2020.
\bibitem{ref31}Mousavi A, Dasarathy G, Baraniuk R G. Deepcodec: Adaptive sensing and recovery via deep convolutional neural networks[J]. arXiv preprint arXiv: 1707.03386, 2017.
\bibitem{ref32}Jiang Du, Xuemei Xie, Chenye Wang, Guangming Shi, Perceptual Compressive Sensing, The First Chinese Conference on Pattern Recognition and Computer Vision (PRCV), November, 2018.
\bibitem{ref33}Du J, Xie X, Wang C, et al. Fully convolutional measurement network for compressive sensing image reconstruction[J]. Neurocomputing, vol.328. pp.105-112, 2019
\end{thebibliography}
\end{document}